\documentclass[10pt,twocolumn,letterpaper]{article}

\usepackage[pagenumbers]{cvpr} 

\definecolor{cvprblue}{rgb}{0.21,0.49,0.74}
\usepackage[pagebackref,breaklinks,colorlinks,allcolors=cvprblue]{hyperref}
\usepackage{threeparttable}
\usepackage{multirow}
\usepackage[normalem]{ulem}
\useunder{\uline}{\ul}{}
\usepackage{booktabs} 
\usepackage{pgfplots}

\def\paperID{} 
\def\confName{CVPR}
\def\confYear{2026}

\title{Physics-Consistent Diffusion for Efficient Fluid Super-Resolution via \\ Multiscale Residual Correction}

\author{Zhihao Li$^{1}$ \and
Shengwei Dong$^{4}$ \and
Chuang Yi$^{4}$ \and
Junxuan Gao$^{4}$ \and
Zhilu Lai$^{1,2}$ \and
Zhiqiang Liu$^{3}$$^*$ \and
Wei Wang$^{1,2}$$^*$ \and
Guangtao Zhang$^{3,4}$\thanks{Corresponding author.} \and
$^1$The Hong Kong University of Science and Technology (Guangzhou), Guangzhou, China \qquad \\
$^2$The Hong Kong University of Science and Technology, Hong Kong SAR, China \qquad \\
$^3$Southern University of Science and Technology, Shenzhen, China \qquad\\
$^4$SandGold AI Research, Guangzhou, China \qquad \\
{\tt\small {zli416@connect.hkust-gz.edu.cn}; {liuzq@sustech.edu.cn}; {weiwcs@ust.hk}; {tao@sgd-ai.com}}
}

\begin{document}
\maketitle
\begin{abstract}
Existing image SR and generic diffusion models transfer poorly to fluid SR: they are sampling-intensive, ignore physical constraints, and often yield spectral mismatch and spurious divergence. 
We address fluid super-resolution (SR) with \textbf{ReMD} (\underline{Re}sidual-\underline{M}ultigrid \underline{D}iffusion), a physics-consistent diffusion framework. 
At each reverse step, ReMD performs a \emph{multigrid residual correction}: the update direction is obtained by coupling data consistency with lightweight physics cues and then correcting the residual across scales; the multiscale hierarchy is instantiated with a \emph{multi-wavelet} basis to capture both large structures and fine vortical details. 
This coarse-to-fine design accelerates convergence and preserves fine structures while remaining equation-free. 
Across atmospheric and oceanic benchmarks, ReMD improves accuracy and spectral fidelity, reduces divergence, and reaches comparable quality with markedly fewer sampling steps than diffusion baselines. 
Our results show that enforcing physics consistency \emph{inside} the diffusion process via multigrid residual correction and multi-wavelet multiscale modeling is an effective route to efficient fluid SR. 
Our code are available on \url{https://github.com/lizhihao2022/ReMD}.
\end{abstract}
    
\begin{figure*}
\centering
\includegraphics[width=1\linewidth]{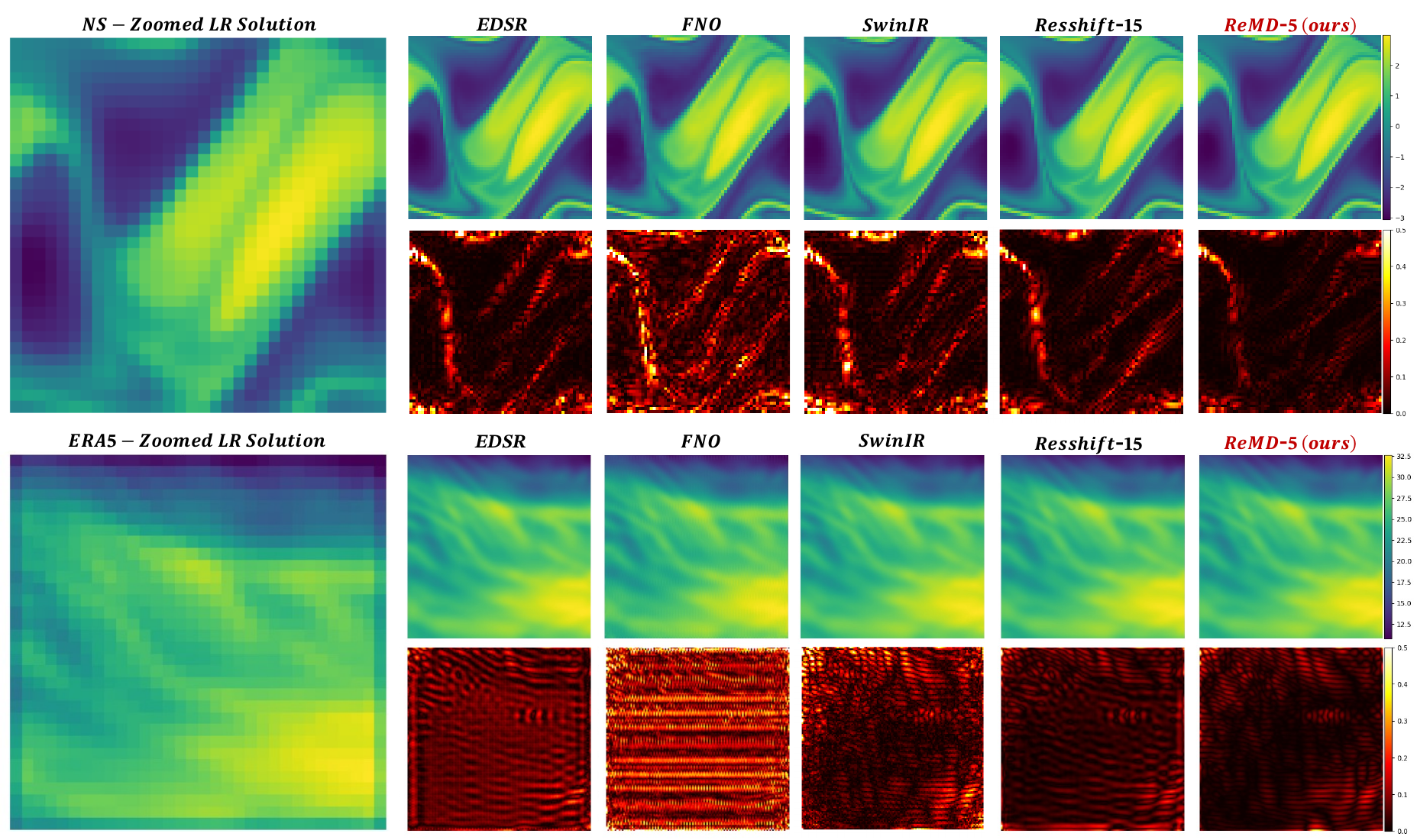}
\vspace{-20pt}
\caption{\textbf{Qualitative comparison on NS (\,$\times$2\,) and ERA5 (\,$\times$4\,).}
Each block shows (left) a zoomed LR input $u_0$, (top row) HR reconstructions, and (bottom row) absolute error maps w.r.t.\ ground truth (shared color scale per block).
With only \textbf{5} reverse steps, \textbf{ReMD} yields (NS) sharper filaments/coherent fronts and (ERA5) preserved mesoscale shear bands while suppressing ringing/stripe artifacts seen in EDSR/FNO/SwinIR, and achieves lower errors than \textit{ResShift} despite requiring \textbf{5} vs.\ \textbf{15} steps.}
\label{fig:vis}
\end{figure*}

\section{Introduction}
\label{sec:intro}

High-resolution (HR) fluid fields are crucial for understanding transport, extremes, and subgrid-scale variability in climate and computational fluid dynamics (CFD). However, operational systems typically run at coarse resolution to satisfy wall-time and cost constraints. Running full-physics solvers or learned surrogates at HR remains one to two orders of magnitude more expensive than at low resolution. As a result, a practical pipeline has emerged: first obtain a coarse solution using a numerical model or a neural operator \cite{li2020fourier,kovachki2023neural}, then apply a learned super-resolution (SR) model to reconstruct HR details \cite{rasp2018deep,stengel2020adversarial,wang2020deep}. 

Directly applying generic image SR to fluid SR is problematic. 
Geophysical and turbulent flows have wide-band, multiscale spectra and filamentary structures whose statistics differ from natural images; pixel/feature losses on RGB data tend to distort both low- and high-frequency content, yielding incorrect energy spectra and missing small eddies. 
Standard SR models also lack physics consistency—often introducing spurious divergence, violating simple flux/boundary behavior, and degrading stability under rollout \cite{kochkov2021machine,brandstetter2022message}. 
Finally, one-shot LR to HR mappings underuse the fact that a coarse solution $u_0$ already approximates the target operator, overlooking the benefits of iterative residual correction familiar from multigrid.

Diffusion models are appealing here because they refine solutions iteratively and can impose strong generative priors for fine structures \cite{saharia2022image,rombach2022high}. 
However, off-the-shelf diffusion for images is ill-matched to fluid SR: the forward process injects pure noise that drifts away from coarse/physical manifolds, the reverse updates are typically oblivious to physics and spectra, and many sampling steps inflate cost. 
Our approach addresses these gaps by placing a multiscale, physics-aware residual correction \emph{inside} the reverse process. 
A preview in Fig.~\ref{fig:vis} shows that, given the same LR initial solution, \textbf{ReMD} reconstructs sharper vortical structures and yields substantially lower error maps than diffusion and non-diffusion baselines, using only \emph{5} reverse steps versus \emph{15} for ResShift.

We instead view fluid SR as \emph{iterative residual correction} on top of a given coarse solution, in the spirit of multigrid methods \cite{wl2000mg,juncai2024mgno,zhihao2024m2no}. At each reverse step, we want to measure how inconsistent the current HR estimate $u_t$ is with the available low-resolution information and with lightweight physics cues, and then correct this residual across scales. This connects to residual-based diffusion such as ResShift \cite{yue2023resshift}, which samples in a residual space rather than directly in image space. Our formulation is similar in spirit but differs in three key aspects. 
(i) The residual is \emph{recomputed at every step} from a data-consistency term and an equation-free physics residual, rather than being a fixed HR--LR difference. 
(ii) The correction is performed by a \emph{multiscale} operator derived from multigrid and multiwavelet ideas.
(iii) The design is tailored to fluid fields: it operates on scalar/vector components, respects simple physical structure (e.g., divergence, masks, spectra), and can be used on top of coarse solutions produced by either classical solvers or neural operators.

Building on this view, we introduce \textbf{ReMD} (\emph{Residual–Multigrid Diffusion}), a physics-consistent diffusion framework for fluid SR. \textbf{ReMD} realized an operator from the coarse space to HR with few, coarse-to-fine steps, improving spectral fidelity and reducing spurious divergence. Our main contributions are:
\begin{itemize}[leftmargin=1.1em]
\item \textbf{Formulation.} We cast fluid SR as operator learning via few-step \emph{residual} diffusion: instead of direct denoising, ReMD iteratively corrects a coarse solution with physics-consistent updates.
\item \textbf{Multiscale drift.} We embed a time-gated multigrid V-cycle (multiwavelet restriction/prolongation with lightweight learned smoothers) as the per-step drift inside the sampler, yielding stable coarse-to-fine corrections.
\item \textbf{Equation-free physics.} We design fully differentiable, inexpensive physics residuals (e.g., divergence, spectrum alignment) that require no PDE at test time.
\item \textbf{Effectiveness and efficiency.} Across NS, ERA5, and Ocean benchmarks, ReMD attains higher accuracy and spectral fidelity with lower divergence than diffusion and non-diffusion baselines, while matching or exceeding their quality with markedly fewer sampling steps.
\end{itemize}

\section{Preliminaries}
\label{sec:prelim}
In this section, we formalize fluid super-resolution as an operator-learning map from coarse fields to high-resolution states and fix the notation used throughout (\S\ref{subsec:formulation}). 
We then recall the classical multigrid residual-correction principle, which later serves as the time-conditioned corrector inside our diffusion refinement (\S\ref{subsec:mg}).

\subsection{Problem formulation}
\label{subsec:formulation}
We cast fluid SR as \emph{operator learning} between function spaces. 
Let $\mathcal{U}$ be the HR state space (e.g., scalar/vector fields on a spatial domain) and $\mathcal{U}_{\text{c}}$ its coarse counterpart.
Given a coarse field $u^{\text{LR}}\!\in\!\mathcal{U}_{\text{c}}$, we seek to reconstruct the unknown HR field $u^{\text{HR}}\!\in\!\mathcal{U}$ by learning
\begin{equation}
\mathcal{F}_\theta:\; \mathcal{U}_{\text{c}} \to \mathcal{U}, 
\qquad 
\hat{u}^{\text{HR}}=\mathcal{F}_\theta\!\left(u^{\text{LR}}\right).
\end{equation}

\paragraph{Restriction consistency.}
We assume a known restriction operator $R:\mathcal{U}\!\to\!\mathcal{U}_{\text{c}}$ (e.g., averaging/downsampling) and aim for coarse-scale consistency $R(\hat{u}^{\text{HR}})\!\approx\!u^{\text{LR}}$.
We do \emph{not} assume access to governing equations or to the true solution operator; $u^{\text{LR}}$ may come from any upstream pipeline, and $R$ provides a common coarse-space reference.

\paragraph{Iterative diffusion SR.}
We realize $\mathcal{F}_\theta$ as a finite reverse-time refinement.
Let $u_t$ be the HR estimate at step $t$ ($t=T,\dots,1,0$). We update
\begin{equation}
u_{t-1} \;=\; u_t \;+\; \alpha_t\, e_t \;+\; \sigma_t\,\varepsilon_t,
\qquad
\varepsilon_t \sim \mathcal{N}(0,\mathbf{I}),
\end{equation}
with schedules $\alpha_t,\sigma_t$. Define the residual and the time–conditioned corrector by
\[
r(u) \;=\; u^{\text{LR}} - R(u), 
\qquad
e_t \;=\; \mathsf{S}_t\!\big(r(u_t)\big),
\]
where $\mathsf{S}_t$ is the residual corrector with timestep gating. Thus,
\[
\mathcal{F}_\theta:\; u^{\text{LR}} \;\mapsto\; u_T \;\mapsto\; \cdots \;\mapsto\; u_0 \;\equiv\; \hat{u}^{\text{HR}}.
\]

\subsection{Residual correction via multigrid}
\label{subsec:mg}
Multigrid (MG) \cite{wl2000mg,jinchao2002method} accelerates iterative solvers by letting each scale remove the errors it sees best: smooth (low–frequency) errors on coarse grids and oscillatory (high–frequency) errors on fine grids. 
Consider a linear system on a fine grid $\Omega^{h}$,
\begin{equation}
A^{h} u^{h} = b^{h},
\end{equation}
with linear restriction $I_{h}^{2h}$ and prolongation $I_{2h}^{h}$ between $\Omega^{h}$ and $\Omega^{2h}$.  
Let $u^{h,(k)}$ be the current fine–grid iterate (iteration index $k$). A two–level V–cycle that maps $u^{h,(k)} \mapsto u^{h,(k+1)}$ proceeds as:

\paragraph{1) Pre-smoothing.}
Damp high-frequency error on the fine grid:
\begin{equation}
\tilde{u}^{h,(k)} \;\leftarrow\; \mathsf{Smooth}^{h}\!\big(u^{h,(k)}, A^{h}, b^{h}\big).
\end{equation}

\paragraph{2) Residual transfer.}
Form the fine-grid residual and restrict it to the coarse grid:
\begin{equation}
r^{h,(k)} \;=\; b^{h} - A^{h}\tilde{u}^{h,(k)}, 
\qquad
r^{2h,(k)} \;=\; I_{h}^{2h}\, r^{h,(k)}.
\end{equation}

\paragraph{3) Coarse correction.}
Approximately solve the coarse error equation:
\begin{equation}
A^{2h} e^{2h,(k)} \;=\; r^{2h,(k)}.
\end{equation}

\paragraph{4) Prolongation and update.}
Lift the coarse error and correct the fine–grid state:
\begin{equation}
\hat{u}^{h,(k)} \;\leftarrow\; \tilde{u}^{h,(k)} + I_{2h}^{h}\, e^{2h,(k)}.
\end{equation}

\paragraph{5) Post-smoothing.}
Remove high-frequency components reintroduced by prolongation and obtain the next iterate:
\begin{equation}
u^{h,(k+1)} \;\leftarrow\; \mathsf{Smooth}^{h}\!\big(\hat{u}^{h,(k)}, A^{h}, b^{h}\big).
\end{equation}

\noindent One V–cycle thus defines an iterative mapping $u^{h,(k+1)}=\mathsf{V\text{-}cycle}\!\big(u^{h,(k)}\big)$.  
Later, we reuse this classical “residual $\rightarrow$ restrict $\rightarrow$ coarse correction $\rightarrow$ prolong $\rightarrow$ smooth” principle as a time–conditioned residual corrector $\mathsf{S}_t$ for constructing the update direction $e_t$ in the iterative diffusion refinement (where the role of $u^{h,(k)}$ is played by the current HR estimate $u_t$).

\section{Methodology}
\label{sec:method}
Building on the operator view in Sec.~\ref{sec:prelim}, we instantiate \textbf{ReMD} as a few-step diffusion sampler in which \emph{each reverse step performs a multiscale residual correction}. Concretely, at every step we (i) form a residual that enforces coarse-scale consistency and lightweight physics cues, (ii) transport that residual across resolutions with a time-gated multigrid pass using \emph{fixed} multiwavelet restriction/prolongation, and (iii) apply a small learned head to make a minor refinement. 

\begin{figure*}[t]
    \centering
    \includegraphics[width=\linewidth]{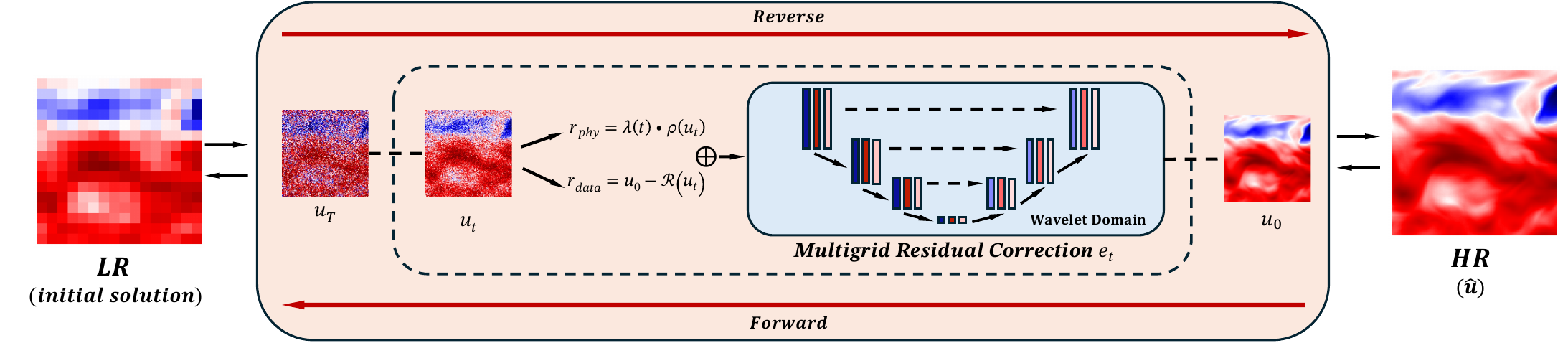}
    \caption{Overview of ReMD. Starting from a coarse LR initial solution, the reverse diffusion steps refine the HR estimate $u_t$ while a multigrid residual correction module combines data-consistency and physics residuals in the wavelet domain, yielding the final HR field $\hat{u}$.}
    \label{fig:overview}
\end{figure*}

The sampler composition is described in Sec.~\ref{subsec:sampler}; the time-gated multigrid corrector and its multiwavelet transfers are detailed in Sec.~\ref{subsec:mgcorr}; the lightweight, equation-free physics cues are specified in Sec.~\ref{subsec:phys}; and training choices and schedules are summarized in Sec.~\ref{subsec:training}. Fig.~\ref{fig:overview} provides a high-level overview of the residual formation, multiscale correction, and reverse update.

\subsection{Sampler with multiscale residual drift}
\label{subsec:sampler}
Let \(u_t\) be the HR estimate at reverse step \(t\) (\(t=T,\ldots,1,0\)).
We update
\begin{equation}
u_{t-1} \;=\; u_t \;+\; \alpha_t\, e_t \;+\; \beta_t\, g_\theta(u_t,t) \;+\; \sigma_t\,\varepsilon_t,
\qquad
\varepsilon_t \sim \mathcal{N}(0,I),
\label{eq:remd_update}
\end{equation}
where \(g_\theta\) is a small learned head (zero-initialized last layer), and \(\alpha_t,\beta_t,\sigma_t\) are schedules (DDIM when \(\sigma_t{=}0\)).
The drift direction \(e_t\) is obtained from a \emph{time-conditioned residual corrector}
\begin{equation}
r(u) \;=\; u^{\text{LR}} - R(u) \;+\; \lambda(t)\,\rho(u),
\qquad
e_t \;=\; S_t\!\big(r(u_t)\big),
\label{eq:remd_residual_drift}
\end{equation}
with known restriction \(R\) (e.g., averaging/downsampling), lightweight physics cues \(\rho(\cdot)\) (Sec.~\ref{subsec:phys}), and a time-gated MG corrector \(S_t\) (Sec.~\ref{subsec:mgcorr}).

\subsection{Time-gated multigrid residual corrector}
\label{subsec:mgcorr}
The corrector \(S_t\) implements the classical “residual \(\rightarrow\) restrict \(\rightarrow\) coarse correction \(\rightarrow\) prolong \(\rightarrow\) smooth” principle in a learnable, \emph{time-gated} form:
\begin{align}
S_t(r)
\;=\;&\; \mathrm{Smooth}_0(r)
\;+\; \sum_{\ell=1}^{L} w_\ell(t)\, P_\ell\, \mathrm{Smooth}_\ell\!\big(R_\ell r\big),
\label{eq:remd_st}
\end{align}
where \(R_\ell,P_\ell\) are fixed multiwavelet restriction/prolongation operators, \(\mathrm{Smooth}_\ell\) are small conv smoothers, and gates \(w_\ell(t)\!\in\!(0,1)\) come from a timestep embedding followed by a tiny MLP and sigmoid. 
Early steps emphasize coarse levels (removing low-frequency errors); later steps emphasize fine levels (refining sharp fronts). 
Because transfers are fixed and depthwise-separable smoothers are used, the per-step complexity is \(\mathcal{O}(HW)\) per level with a small constant.

\paragraph{Multiwavelet mapping operators.}
Inspired by M2NO \cite{zhihao2024m2no}, we implement the multigrid residual corrector with \emph{fixed multiwavelet transfers} so that restriction/prolongation are spectrally clean and parameter–free, while only within–level smoothers are learned. Let \(H_x,H_y\) be 1D low–pass (scaling) filters from an orthonormal multiwavelet system; in 2D we use separable tensors
\begin{equation}
R \;\equiv\; I_h^{2h} \;=\; H_y \otimes H_x,
\qquad
P \;\equiv\; I_{2h}^{h} \;=\; R^{\!\top},
\end{equation}
where $\otimes$ is the Kronecker product. On a hierarchy, level–wise filters \(R_\ell,P_\ell\) are obtained by dyadic dilation/shift of the base filters; we keep them \emph{fixed}. This yields stable, \(\mathcal{O}(HW)\) inter–level mappings with sharp low/high–frequency separation, while the learned smoothers are tiny depthwise \(3{\times}3\) convolutions gated by the timestep.

Intuitively, coarse levels remove large–scale bias early; fine levels sharpen fronts/vortices later. Because \(R_\ell,P_\ell\) are fixed and smoothers are depthwise, the per–step cost stays close to standard CNN diffusion.

\begin{table*}[t]
\caption{\textbf{Quantitative comparison on three fluid SR benchmarks.} NS ($2{\times}$), ERA5 ($4{\times}$), and Ocean ($4{\times}$). We report RMSE$\downarrow$, PSNR$\uparrow$, and SSIM$\uparrow$. \textbf{Bold} denotes the best and \underline{underline} the second-best within each dataset/metric block. \textbf{ReMD} attains the lowest errors and top (or tied) perceptual scores across datasets while using only 2–5 reverse steps, compared with 15 steps for the ResShift baseline.}
\vspace{-8pt}
\label{table:main}
\begin{sc}
\renewcommand{\multirowsetup}{\centering}
\footnotesize 
\resizebox{\linewidth}{!}{
\begin{tabular}{c|ccc|ccc|ccc}
\toprule
\multirow{2}{*}{Model} & \multicolumn{3}{c}{NS} & \multicolumn{3}{c}{ERA5} & \multicolumn{3}{c}{Ocean} \\ 
\cmidrule(lr){2-4} \cmidrule(lr){5-7} \cmidrule(lr){8-10} & RMSE $\downarrow$ & PSNR $\uparrow$ & SSIM $\uparrow$ & RMSE $\downarrow$ & PSNR $\uparrow$ & SSIM $\uparrow$ & RMSE $\downarrow$ & PSNR $\uparrow$ & SSIM $\uparrow$ \\
\midrule
EDSR            & 2.97E-02 & 46.89 & 0.996 & 9.06E-02 & 57.13 & \underline{0.998} & 1.36E-02 & 47.48 & \underline{0.982} \\
\midrule
FNO             & 4.42E-02 & 43.44 & 0.988 & 1.35E-01 & 53.69 & 0.997 & 1.67E-02 & 45.74 & 0.974\\
MWT             & 8.45E-02 & 37.81 & 0.970 & 6.13E-01 & 40.53 & 0.968 & 4.55E-02 & 37.03 & 0.852\\
HiNOTE          & 8.03E-02 & 38.25 & 0.973 & 2.12E-01 & 49.76 & 0.994 & 1.99E-02 & 44.21 & 0.964\\
\midrule
Galerkin        & 5.32E-02 & 41.83 & 0.981 & 3.45E-01 & 45.51 & 0.988 & 3.75E-02 & 38.70 & 0.890\\
SwinIR          & 3.63E-02 & 45.15 & 0.994 & 9.10E-02 & 57.09 & \underline{0.998} & 1.35E-02 & 47.55 & \textbf{0.983}\\
\midrule
SR3             & 3.34E-01 & 25.87 & 0.841 & 7.13E+00 & 17.37 & 0.762 & 7.86E-02 & 32.27 & 0.888\\
Resshift-15     & 2.21E-02 & 49.47 & \underline{0.997} & 8.79E-02 & 57.39 & \underline{0.998} & 1.36E-02 & 47.50 & 0.981\\
\midrule
\textbf{ReMD-2} & \underline{2.11E-02} & \underline{49.84} & \textbf{0.998} & \underline{8.03E-02} & \underline{58.13} & \textbf{0.999} & \textbf{1.32E-02} & \textbf{47.72} & \textbf{0.983}\\
\textbf{ReMD-5} & \textbf{2.09E-02} & \textbf{49.94} & \textbf{0.998} & \textbf{8.02E-02} & \textbf{58.19} & \textbf{0.999} & \underline{1.33E-02} & \underline{47.71} & \textbf{0.983}\\
\bottomrule
\end{tabular}}
\end{sc}
\end{table*}

\subsection{Physics-consistent residuals}
\label{subsec:phys}

We regularize the reverse updates with lightweight, differentiable \emph{physics-consistent residuals} that operate on a \emph{single} scalar field (e.g., $u$, $u_x$, $v_x$, temperature). Each residual returns a pixel-space direction with the same shape as the input and is fully backpropagable. Let $u\!\in\!\mathbb{R}^{1\times H\times W}$ be the target field at step $t$, $u_0$ its coarse/anchor field, and $M\!\in\!\{0,1\}^{1\times H\times W}$ an optional fluid mask ($1$=fluid). We write the total physics residual as
\begin{equation}
\label{eq:phys_total}
\rho(u)
\;=\;
\sum_{k} \, w_k \,\rho_k(u; u_0, M),
\end{equation}
where $r_{\text{data}}(u)$ is the data-consistency residual (e.g., $u_0\!-\!R(u)$), $w_k$ are fixed weights, and $\lambda(t)$ is a schedule.

\vspace{4pt}\noindent\textbf{(1) Laplacian / Biharmonic smoothing.}
To suppress spurious oscillations while preserving large scales, we use the negative gradients of quadratic smoothness energies:
\begin{equation}
\rho_{\text{lap}}(u) \;=\; -\Delta u,
\qquad
\rho_{\text{bi}}(u) \;=\; -\Delta(\Delta u),
\end{equation}
implemented with small-depthwise convolutions (stable and fast). $\rho_{\text{bi}}$ more aggressively removes checkerboard/ringing artifacts and is useful for derivative fields such as $u_x$.

\vspace{4pt}\noindent\textbf{(2) Anisotropic edge-preserving diffusion.}
We protect sharp fronts/filaments using a Perona--Malik style flux guided by an anchor (coarse) field $u_a$ (default $u_a\!=\!u_0$):
\begin{equation}
\rho_{\text{aniso}}(u)
\;=\;
-\nabla\!\cdot\!\Big(g\!\big(\|\nabla u_a\|\big)\,\nabla u\Big),\quad
g(s)=\frac{1}{1+(s/\kappa)^2}.
\end{equation}
Early steps emphasize coarse guidance; later steps are relaxed via time gating inside $\rho_{\text{aniso}}$.

\vspace{4pt}\noindent\textbf{(3) Spectrum alignment.}
We match the radial log-power spectrum of $u$ to an anchor (e.g., $u_0$) to enforce realistic spectral slopes while remaining equation-free. Let $\mathcal{F}$ be the FFT and $\mathcal{B}$ the inverse FFT. With bin-wise weights $W(k)$ derived from binned log-power discrepancies,
\begin{equation}
\rho_{\text{spec}}(u)
\;=\;
\mathcal{B}\!\big(W(k)\odot \mathcal{F}(u)\big),
\end{equation}
where $W(k)$ uses a robust (Huber) transform of $\log P(u)$ minus the target log-power. Optional masking reduces coastline-induced ringing.

We combine the above residuals as in Eq.~\eqref{eq:phys_total}. Each term is implemented via depthwise convolutions or FFT/iFFT, thus fully differentiable and efficient ($\mathcal{O}(HW)$ per pass). A cosine $\lambda(t)$ emphasizes physics early and decays moderately later, while fine-grained time gating inside individual $\rho_k$ (e.g., in $\rho_{\text{aniso}}$) produces a coarse-to-fine prior that complements the multiscale residual correction in the reverse process.

\subsection{Training objective and schedules}
\label{subsec:training}
We adopt standard diffusion training with an \(\varepsilon\)-prediction loss and a cosine noise schedule; the MG drift is used in the reverse mean during training and inference via Eq.~\eqref{eq:remd_update}–\eqref{eq:remd_residual_drift}.
Given forward-diffused pairs \((u_t,\varepsilon)\),
\begin{equation}
\mathcal{L}(\theta)
\;=\;
\mathbb{E}_{t,u_t,\varepsilon}\!
\Big[\;\|\varepsilon-\hat{\varepsilon}_\theta(u_t,t)\|_2^2\;\Big],
\end{equation}
and we instantiate the reverse mean as
\[
\mu_\theta(u_t,t)
= u_t + \alpha_t\, e_t + \beta_t\, g_\theta(u_t,t),
\quad e_t=S_t\!\big(r(u_t)\big).
\]
At test time, we use DDIM (\(\sigma_t{=}0\)) with \(5\!-\!10\) NFEs; by keeping the trajectory close to the restriction/physics manifold, ReMD attains target quality with markedly fewer steps than vanilla diffusion.
\section{Experiments}
\label{sec:exp}

In this section we evaluate \textbf{ReMD} under a unified protocol: we outline datasets, training and metrics, then report main results with representative visual comparisons. We further probe behaviour via frequency–domain analysis, patch-level spatial inspection, and ablations that isolate the roles of multigrid correction and physics cues, followed by brief discussion of design implications.

\definecolor{myred}{RGB}{219,049,036}
\definecolor{myorange}{RGB}{252,140,090}
\definecolor{myblue}{RGB}{075,116,178}
\definecolor{mywater}{RGB}{144,190,224}
\definecolor{myyellow}{RGB}{255,223,146}
\definecolor{mygrey}{RGB}{230,241,243}

\definecolor{vir0}{RGB}{068,004,090}
\definecolor{vir1}{RGB}{065,062,133}
\definecolor{vir2}{RGB}{048,104,141}
\definecolor{vir3}{RGB}{031,146,139}
\definecolor{vir4}{RGB}{053,183,119}
\definecolor{vir5}{RGB}{145,213,066}
\definecolor{vir6}{RGB}{248,230,032}

\pgfplotsset{width=0.9\linewidth,height=4cm,scale only axis}
\pgfplotsset{every axis/.append style={
font=\small,
line width=1.0pt,
tick style={line width=0.3pt}}}
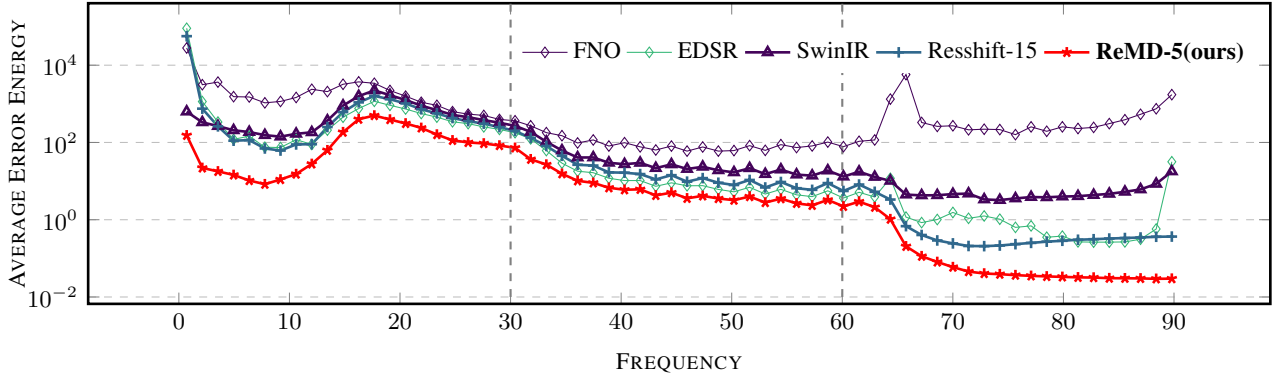
\begin{figure*}[ht]
    \centering
    \begin{tikzpicture}
    \begin{axis}[
    legend style={at={(0.99,0.91)},anchor=north east,legend columns=-1, draw=none},
    xlabel=\textsc{Frequency},
    ylabel=\textsc{Average Error Energy},
    ylabel style={yshift=-0.8em},
    xtick={0,10,20,30,40,50,60,70,80,90},
    ymajorgrids=true,
    ymode=log,
    grid style=dashed,
    colormap={greenyellow}{rgb255(0cm)=(0,128,0); rgb255(1cm)=(255,255,0)},
    ]
\addplot[thin, mark=diamond, color=vir0] coordinates {
(0.7071067690849304, 27881.0456013459)
(2.1213202476501465, 3129.0658329284192)
(3.535533905029297, 3653.4998209340806)
(4.949747085571289, 1533.3988299894634)
(6.363961219787598, 1504.3600629094074)
(7.77817440032959, 1061.05837664477)
(9.192388534545898, 1153.55533093799)
(10.60660171508789, 1461.4271835349084)
(12.020814895629883, 2380.847261419305)
(13.435028076171875, 2073.2746725979496)
(14.849242210388184, 3212.4127649960296)
(16.26345443725586, 3701.5601466168346)
(17.677669525146484, 3408.3760202407943)
(19.091882705688477, 2213.2617022178306)
(20.50609588623047, 1580.7405594101906)
(21.920310974121094, 1097.864519859417)
(23.334522247314453, 914.8443335006192)
(24.748737335205078, 627.7659321159275)
(26.16295051574707, 541.504579530958)
(27.577163696289062, 501.2345041210572)
(28.991378784179688, 394.83941503040495)
(30.405590057373047, 364.05123892053257)
(31.819805145263672, 269.8549716209321)
(33.23401641845703, 179.63475706108812)
(34.648231506347656, 149.6486528465953)
(36.06244659423828, 97.05830021471488)
(37.47665786743164, 116.22714572968466)
(38.890872955322266, 80.8327409935481)
(40.305084228515625, 98.73786152837445)
(41.71929931640625, 77.0147945038276)
(43.133514404296875, 63.80272577230667)
(44.547725677490234, 80.678668101502)
(45.96194076538086, 59.68997493272907)
(47.37615203857422, 76.40690471125218)
(48.790367126464844, 59.771615174776244)
(50.20458221435547, 61.633631659225536)
(51.61879348754883, 82.72444253250063)
(53.03300857543945, 62.484085340307445)
(54.44721984863281, 87.71729032892614)
(55.86143493652344, 73.87643043229697)
(57.27565002441406, 81.28629252267606)
(58.68986129760742, 102.3276592639529)
(60.10407638549805, 77.82163080271545)
(61.518287658691406, 107.18441057822916)
(62.93250274658203, 116.04443744343524)
(64.34671783447266, 1324.1626982742455)
(65.76092529296875, 5710.858996486581)
(67.17514038085938, 329.0015383981839)
(68.58935546875, 260.76322317931255)
(70.00357055664062, 270.4630866268283)
(71.41778564453125, 212.34170985068067)
(72.83200073242188, 220.32954284887154)
(74.24620819091797, 215.52969645733953)
(75.6604232788086, 159.43807857953286)
(77.07463836669922, 253.42657627041805)
(78.48885345458984, 194.69923847849944)
(79.90306091308594, 253.68391833219354)
(81.31727600097656, 229.75496700414067)
(82.73149108886719, 245.66081067274573)
(84.14570617675781, 307.4785730197293)
(85.55992126464844, 387.2393168312533)
(86.97413635253906, 535.1366737665135)
(88.38834381103516, 745.1357401311786)
(89.80255889892578, 1728.096657592884)
};
\addplot[thin, mark=diamond,color=vir4] coordinates {
(0.7071067690849304, 90313.91925170677)
(2.1213202476501465, 1144.832418423839)
(3.535533905029297, 337.4962410649145)
(4.949747085571289, 120.47216824914254)
(6.363961219787598, 145.97349155659945)
(7.77817440032959, 75.34484479510531)
(9.192388534545898, 73.81796554323721)
(10.60660171508789, 118.33141170246925)
(12.020814895629883, 88.915357473988)
(13.435028076171875, 208.52033189795528)
(14.849242210388184, 451.0651854411833)
(16.26345443725586, 755.7874490335123)
(17.677669525146484, 1162.8681702080474)
(19.091882705688477, 916.8196516322352)
(20.50609588623047, 734.4446957107683)
(21.920310974121094, 561.2522935680566)
(23.334522247314453, 452.2680898012035)
(24.748737335205078, 343.1543171277438)
(26.16295051574707, 306.8305212008893)
(27.577163696289062, 253.32072679503593)
(28.991378784179688, 218.04309487142184)
(30.405590057373047, 178.90575927880616)
(31.819805145263672, 124.07524902177944)
(33.23401641845703, 61.58607706134979)
(34.648231506347656, 29.13472165926776)
(36.06244659423828, 18.020324067984632)
(37.47665786743164, 16.496486402078103)
(38.890872955322266, 11.822826830787901)
(40.305084228515625, 10.324511696256517)
(41.71929931640625, 10.329091384237789)
(43.133514404296875, 7.335990916143664)
(44.547725677490234, 9.007057526718773)
(45.96194076538086, 7.568605320800461)
(47.37615203857422, 7.61843402193483)
(48.790367126464844, 6.092844261133949)
(50.20458221435547, 5.277738938628427)
(51.61879348754883, 6.908735622076752)
(53.03300857543945, 4.622264553342632)
(54.44721984863281, 6.117327441308885)
(55.86143493652344, 4.410644504681986)
(57.27565002441406, 3.9609344846285416)
(58.68986129760742, 5.603239512902058)
(60.10407638549805, 3.6421182297712536)
(61.518287658691406, 5.13429500830526)
(62.93250274658203, 3.8827106416691705)
(64.34671783447266, 11.845079422926338)
(65.76092529296875, 1.1695918973801578)
(67.17514038085938, 0.8430003662123629)
(68.58935546875, 1.0124909013264006)
(70.00357055664062, 1.5429958635914496)
(71.41778564453125, 1.0877048225380218)
(72.83200073242188, 1.2519261412503762)
(74.24620819091797, 1.0314889775378136)
(75.6604232788086, 0.6341755548206817)
(77.07463836669922, 0.6929527611802736)
(78.48885345458984, 0.3526082104256933)
(79.90306091308594, 0.3769611844136771)
(81.31727600097656, 0.2668848294572564)
(82.73149108886719, 0.26279471846396973)
(84.14570617675781, 0.2627387648783972)
(85.55992126464844, 0.2665061281809186)
(86.97413635253906, 0.31329659314969545)
(88.38834381103516, 0.5752613280920818)
(89.80255889892578, 31.63650732096393)};
\addplot[mark=triangle,color=vir0] coordinates{
(0.7071067690849304, 628.415111821699)
(2.1213202476501465, 332.40516633901885)
(3.535533905029297, 265.28271542850217)
(4.949747085571289, 206.67228576037064)
(6.363961219787598, 186.16160264405528)
(7.77817440032959, 154.70023873034264)
(9.192388534545898, 141.47830600200187)
(10.60660171508789, 167.4249866612538)
(12.020814895629883, 183.16276248024552)
(13.435028076171875, 368.4313930520456)
(14.849242210388184, 899.2162034582249)
(16.26345443725586, 1581.4213855301759)
(17.677669525146484, 2224.4907122527496)
(19.091882705688477, 1676.326619214079)
(20.50609588623047, 1255.9551502929942)
(21.920310974121094, 899.035852855792)
(23.334522247314453, 688.7618689412168)
(24.748737335205078, 516.5546044477068)
(26.16295051574707, 451.375145834673)
(27.577163696289062, 379.88925470252996)
(28.991378784179688, 323.63854732438374)
(30.405590057373047, 269.24921863779395)
(31.819805145263672, 187.67853482827652)
(33.23401641845703, 104.27962064320171)
(34.648231506347656, 62.03267461570517)
(36.06244659423828, 40.80727549505798)
(37.47665786743164, 40.912537785593386)
(38.890872955322266, 29.784227179820064)
(40.305084228515625, 27.375146781764624)
(41.71929931640625, 29.63905264646463)
(43.133514404296875, 21.69148890222235)
(44.547725677490234, 27.090295186573134)
(45.96194076538086, 20.58126059046985)
(47.37615203857422, 23.45222043461365)
(48.790367126464844, 19.03956910698068)
(50.20458221435547, 16.831592473649907)
(51.61879348754883, 21.548184949957694)
(53.03300857543945, 15.171214019418533)
(54.44721984863281, 19.867492018616346)
(55.86143493652344, 14.709774416061715)
(57.27565002441406, 13.8469724348956)
(58.68986129760742, 18.743058355856226)
(60.10407638549805, 13.046279359853411)
(61.518287658691406, 17.498912806346824)
(62.93250274658203, 12.81468791613873)
(64.34671783447266, 10.149622152637508)
(65.76092529296875, 4.468952132439323)
(67.17514038085938, 4.271215869135973)
(68.58935546875, 4.306126761318774)
(70.00357055664062, 4.618095120420269)
(71.41778564453125, 4.747253217158507)
(72.83200073242188, 3.35074983804409)
(74.24620819091797, 3.227404235290086)
(75.6604232788086, 3.558635204959874)
(77.07463836669922, 3.86253153278867)
(78.48885345458984, 3.809158332141829)
(79.90306091308594, 3.980369749805435)
(81.31727600097656, 4.068519482572907)
(82.73149108886719, 4.362363297065292)
(84.14570617675781, 4.676350342844151)
(85.55992126464844, 5.282319012600766)
(86.97413635253906, 6.085337630271998)
(88.38834381103516, 8.474175926967409)
(89.80255889892578, 17.8156374061646)
};
\addplot[mark=+,color=vir2] coordinates{
(0.7071067690849304, 56399.26371217059)
(2.1213202476501465, 750.1517805832485)
(3.535533905029297, 261.10824453390006)
(4.949747085571289, 111.05139429127794)
(6.363961219787598, 115.02994510901877)
(7.77817440032959, 69.95270991181565)
(9.192388534545898, 61.776031433284906)
(10.60660171508789, 88.87718534874371)
(12.020814895629883, 90.30268227128923)
(13.435028076171875, 250.8992064165714)
(14.849242210388184, 610.3060142070019)
(16.26345443725586, 1090.4024212600993)
(17.677669525146484, 1604.437875665256)
(19.091882705688477, 1312.9131354738367)
(20.50609588623047, 1023.2586473896428)
(21.920310974121094, 722.5535285317054)
(23.334522247314453, 565.393827711541)
(24.748737335205078, 427.6258131698326)
(26.16295051574707, 356.7141253043119)
(27.577163696289062, 306.36403581077275)
(28.991378784179688, 249.5008921199942)
(30.405590057373047, 198.86963501979483)
(31.819805145263672, 131.5577069802266)
(33.23401641845703, 81.17790759827326)
(34.648231506347656, 45.19264668260097)
(36.06244659423828, 26.46253694051167)
(37.47665786743164, 25.2409516008358)
(38.890872955322266, 16.837680118691022)
(40.305084228515625, 16.49220869184895)
(41.71929931640625, 15.191600697464263)
(43.133514404296875, 11.061627092975506)
(44.547725677490234, 14.225813065002999)
(45.96194076538086, 9.671638328197439)
(47.37615203857422, 11.940846668915235)
(48.790367126464844, 9.067023107902813)
(50.20458221435547, 7.923480218304152)
(51.61879348754883, 10.703924484148807)
(53.03300857543945, 6.836865886945787)
(54.44721984863281, 9.486873408127787)
(55.86143493652344, 6.542605303475588)
(57.27565002441406, 5.935641218338984)
(58.68986129760742, 8.881627020063432)
(60.10407638549805, 5.5474493897977)
(61.518287658691406, 8.148505645462253)
(62.93250274658203, 5.2840747032840865)
(64.34671783447266, 3.3294713441116826)
(65.76092529296875, 0.6816164314596048)
(67.17514038085938, 0.40244588956792504)
(68.58935546875, 0.290725151036538)
(70.00357055664062, 0.24402533015801386)
(71.41778564453125, 0.2088529918806574)
(72.83200073242188, 0.20707862071535987)
(74.24620819091797, 0.21479956300752)
(75.6604232788086, 0.23296921068060356)
(77.07463836669922, 0.2514718924474665)
(78.48885345458984, 0.2703675382019229)
(79.90306091308594, 0.2851261461537023)
(81.31727600097656, 0.30319785629876433)
(82.73149108886719, 0.312957705287811)
(84.14570617675781, 0.3255259826839081)
(85.55992126464844, 0.33604835530385185)
(86.97413635253906, 0.3466964167490106)
(88.38834381103516, 0.36227589452913356)
(89.80255889892578, 0.36631966239093544)
};
\addplot[mark=star,color=red] coordinates{
(0.7071067690849304, 152.9470053137839)
(2.1213202476501465, 21.923684460029225)
(3.535533905029297, 17.833425950837225)
(4.949747085571289, 14.445913512133058)
(6.363961219787598, 10.351357374907869)
(7.77817440032959, 8.30572151119068)
(9.192388534545898, 11.002424244759003)
(10.60660171508789, 15.19994948181436)
(12.020814895629883, 27.8280332818459)
(13.435028076171875, 62.874854560169125)
(14.849242210388184, 184.63097669771885)
(16.26345443725586, 399.63369915898445)
(17.677669525146484, 497.96532411731515)
(19.091882705688477, 392.25706361241975)
(20.50609588623047, 310.55403363825965)
(21.920310974121094, 238.26132516707935)
(23.334522247314453, 160.2230452046706)
(24.748737335205078, 113.39452146564155)
(26.16295051574707, 100.22699092707507)
(27.577163696289062, 94.18987295433994)
(28.991378784179688, 83.37007269236085)
(30.405590057373047, 72.65581675072661)
(31.819805145263672, 36.740989321068185)
(33.23401641845703, 26.74297854173105)
(34.648231506347656, 15.496997200031365)
(36.06244659423828, 10.17751406575912)
(37.47665786743164, 9.002256727452895)
(38.890872955322266, 6.7196521059366835)
(40.305084228515625, 5.966457434586378)
(41.71929931640625, 6.1194222582302215)
(43.133514404296875, 4.261515442681974)
(44.547725677490234, 5.0232236668610425)
(45.96194076538086, 3.5778382772500165)
(47.37615203857422, 4.1454099779119315)
(48.790367126464844, 3.5400927400836793)
(50.20458221435547, 3.2101535695600656)
(51.61879348754883, 3.9789738951268645)
(53.03300857543945, 2.789358409879972)
(54.44721984863281, 3.5051869515550865)
(55.86143493652344, 2.6535576891869176)
(57.27565002441406, 2.356514304566868)
(58.68986129760742, 3.257191775928697)
(60.10407638549805, 2.2130248489906704)
(61.518287658691406, 2.9373234941924184)
(62.93250274658203, 2.1105494763436866)
(64.34671783447266, 1.0336879631762557)
(65.76092529296875, 0.20485420648022215)
(67.17514038085938, 0.11376911263088949)
(68.58935546875, 0.08067044643920329)
(70.00357055664062, 0.059323749026638165)
(71.41778564453125, 0.0454940431876277)
(72.83200073242188, 0.04053613652890408)
(74.24620819091797, 0.038946465336935126)
(75.6604232788086, 0.036772591632815225)
(77.07463836669922, 0.035242720053380895)
(78.48885345458984, 0.03407013665332178)
(79.90306091308594, 0.03315323892583912)
(81.31727600097656, 0.032252461596953816)
(82.73149108886719, 0.03181152472150152)
(84.14570617675781, 0.03070335245354584)
(85.55992126464844, 0.030518853344762658)
(86.97413635253906, 0.03026779864166657)
(88.38834381103516, 0.02954080696024318)
(89.80255889892578, 0.029836717217220983)
};
\draw[dashed, thick, gray] (axis cs:30,1e-2) -- (axis cs:30,1e6);
\draw[dashed, thick, gray] (axis cs:60,1e-2) -- (axis cs:60,1e6);
\legend{FNO,EDSR,SwinIR,Resshift-15,\textbf{ReMD-5(ours)}}
\end{axis}
\end{tikzpicture}
\vspace{-10pt}
\caption{\textbf{Error–energy spectrum on ERA5 (,$\times$4,).} Radial average of the Fourier‐domain error (log scale on y) versus frequency (x). Vertical dashed lines mark the LR Nyquist band and the transition toward the HR band. \textbf{ReMD-5} (red) maintains the lowest error from large to high scales, remaining below \textit{ResShift-15}, FNO and image-SR baselines (EDSR, SwinIR), indicating superior spectral fidelity.}
\label{fig:spectral_era5}
\end{figure*}

\subsection{Experimental Setup}
\paragraph{Testing Datasets.}
We evaluate on three benchmarks: Navier–Stokes (NS) flows from PDEBench~\cite{takamoto2022pdebench}, ERA5 reanalysis~\cite{hersbach2023era5}, and the Global Ocean Surface Velocity dataset~\cite{cmems2023global}.
The SR settings are fixed per dataset: \textbf{NS} uses $2{\times}$ SR; \textbf{ERA5} and \textbf{Ocean} use $4{\times}$ SR.
For ERA5 and Ocean we extract HR patches of $256{\times}256$ via sliding windows; the corresponding LR inputs are generated by applying the RealESRGAN degradation operator~\cite{wang2021real} with the prescribed scale. NS follows the dataset’s native grid and the same degradation protocol for the $2{\times}$ setting.

\paragraph{Training Details.}
Unless noted otherwise, we train a single ReMD model per dataset/scale using Adam~\cite{diederik2014adam} (PyTorch~\cite{paszke2019pytorch}) with batch size $64$, learning rate $5\times10^{-5}$, and $\sim$100k iterations. LR–HR pairs are formed using the RealESRGAN degradation at the corresponding scale ($2{\times}$ for NS; $4{\times}$ for ERA5/Ocean). Additional dataset statistics and preprocessing details are provided in the appendix.

\paragraph{Compared Methods.}
We evaluate the effectiveness of ReMD in comparison to nine recent methods, including two diffusion-based models, SR3~\cite{zheng2018unified} and ResShift~\cite{yue2023resshift}, one CNN-based method, EDSR~\cite{kuriakose2023edsr}, two transformer-based methods, SwinIR~\cite{liang2021swinir} and Galerkin Transformer~\cite{cao2021choose}, as well as three established operator learning methods, namely Fourier Neural Operator (FNO)~\cite{li2020fourier}, Multiwavelet Transform (MWT)~\cite{gupta2021multiwavelet}, and HiNOTE~\cite{luo2024hierarchical}.

\paragraph{Metrics.}
We report \textbf{RMSE} (primary), \textbf{PSNR}, and \textbf{SSIM}.
To evaluate \emph{physics-related proxies}, we additionally report
(i) \textbf{VE/EE} on NS2D and (ii) \textbf{GED} on ERA5\_uo, and we analyze
\textbf{radial spectra} (energy / error-energy) to quantify frequency fidelity.
These proxies provide lightweight diagnostics of physically plausible structure,
but they do \emph{not} guarantee satisfaction of the governing PDE.

\subsection{Experimental Results}
\label{subsec:results}

We evaluate on three settings, each with a single upscaling factor: \textbf{NS} ($2{\times}$), \textbf{ERA5} ($4{\times}$), and \textbf{Ocean} ($4{\times}$). Tab.~\ref{table:main} reports RMSE/PSNR/SSIM; Fig.~\ref{fig:vis} provides qualitative comparisons (NS on top, ERA5-$u$ on bottom).

\paragraph{NS (synthetic).}
\textbf{ReMD} attains the best overall scores, surpassing image SR models (EDSR, SwinIR), neural operators (FNO, MWT, HiNOTE), and diffusion baselines (SR3, ResShift). Notably, even \emph{ReMD-2} (two reverse steps) rivals or exceeds stronger baselines, while \emph{ReMD-5} gives a small additional gain. The error maps in Fig.~\ref{fig:vis} (top) show sharper filaments and reduced small-scale errors compared to ResShift (15 steps) and non-diffusion SR, supporting our design of few-step, multiscale residual correction.

\paragraph{ERA5 (reanalysis).}
On realistic reanalysis fields, \textbf{ReMD} matches or improves upon the strongest image SR baselines in perceptual fidelity (PSNR/SSIM) while \emph{also} lowering RMSE. The qualitative example (Fig.~\ref{fig:vis}, bottom) highlights cleaner jets/fronts and fewer artifacts in the error maps. Neural-operator baselines degrade more noticeably on this dataset, whereas ReMD maintains spectral fidelity and coherence—evidence that time-gated multigrid correction plus lightweight physics cues transfers beyond synthetic flows.

\paragraph{\textsc{Ocean} (reanalysis).}
\textbf{ReMD} delivers the lowest numerical error and top or tied perceptual metrics. Competing methods either oversmooth mesoscale structures or introduce high-frequency noise, while ReMD preserves coherent, fine-scale patterns without sacrificing coarse-scale consistency, aligning with our operator view and restriction-based residual.

\paragraph{Summary.}
Across all three benchmarks, \textbf{ReMD} consistently improves RMSE while achieving state-of-the-art or tied PSNR/SSIM—\emph{with only 2--5 reverse steps}, versus 15 for a strong diffusion baseline. These results support our motivation and design: (i) treating SR as iterative \emph{residual} correction anchored by restriction consistency, and (ii) using a \emph{time-gated multigrid} corrector to remove large-scale bias early and refine high-frequency structure later.

\begin{figure*}
\centering
\includegraphics[width=\linewidth]{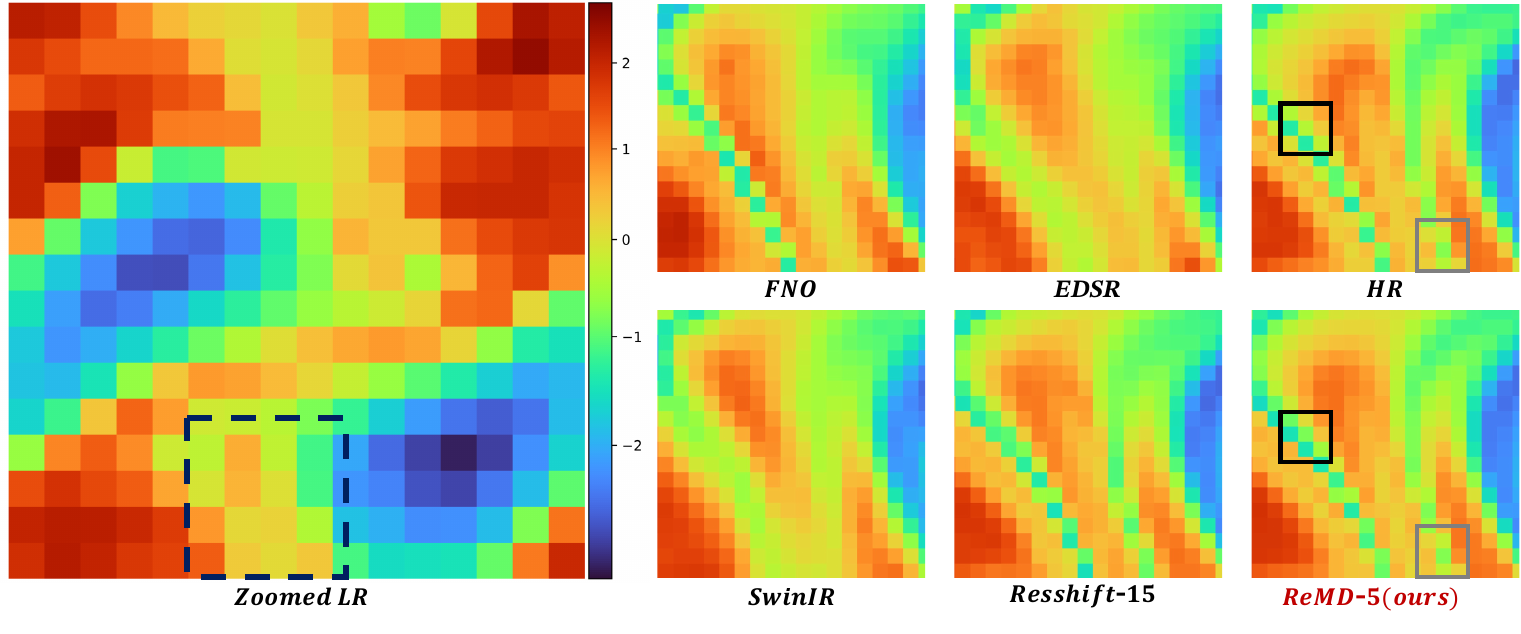}
\vspace{-22pt}
\caption{\textbf{Patch-level comparison on NS ($\times$4).} Left: zoomed LR input. Black boxes highlight frontal bands; gray boxes mark small eddies. \textbf{ReMD–5} recovers sharper, HR-like fronts and coherent vortices with fewer steps than ResShift–15, while avoiding texture/aliasing artifacts seen in image-SR baselines and blockiness in FNO, and remaining consistent with the LR content.}
\label{fig:ns_patch}
\end{figure*}

\subsection{Model Analysis}
\label{subsec:model_analysis}

\subsubsection{Frequency-Domain Behaviour}
\label{subsec:model_spectral}
We examine the radial error-energy spectrum on ERA5 (Fig.~\ref{fig:spectral_era5}) to understand behaviour beyond pixel metrics. The two vertical dashed lines mark (i) the LR Nyquist band where large-scale flow energy concentrates, and (ii) the transition to very high wavenumbers near the HR Nyquist.

\paragraph{Large scales (left of first dash).}
ReMD exhibits the lowest error among all methods, indicating that its coarse-to-fine residual correction reduces large-scale bias rather than merely sharpening textures. Other methods show noticeably larger low-frequency errors, consistent with the visual mismatch of broad structures in Fig.~\ref{fig:vis}.

\paragraph{Mid scales (between dashes).}
This band requires synthesizing subgrid content not present in the LR input. ReMD maintains a uniformly lower curve, reflecting the effect of the time-gated multigrid corrector that removes remaining smooth errors on coarse levels and progressively refines fronts/filaments on finer levels.

\paragraph{High scales (right of second dash).}
Near the HR Nyquist, ReMD keeps the error one to two orders lower than baselines, whereas SwinIR remains elevated and EDSR shows a late spike (ringing/aliasing). The spectrum-faithful tail aligns with our physics-consistent residuals and fixed multiwavelet transfers, which avoid hallucinated high-frequency textures.

\noindent\textit{Summary.} Across scales, ReMD delivers the most spectrally faithful reconstructions, explaining its gains in metrics and the cleaner error maps shown in Fig.~\ref{fig:vis}.

\subsubsection{Patch-level spatial analysis}
\label{subsec:model_spatial}
Fig.~\ref{fig:ns_patch} contrasts a zoomed LR input with HR and baselines (FNO, EDSR, SwinIR, ResShift-15) versus \textbf{ReMD-5}. Key observations:

\begin{itemize}
\item \textbf{Front coherence.} Along the oblique front (black boxes), \textbf{ReMD} preserves sharp cross–front gradients and smooth along–front variation, closely matching HR; EDSR/SwinIR oversmooth and FNO staircases. ResShift-15 sharpens but shows haloing/misalignment.
\item \textbf{Fine-scale eddy.} In the corner eddy (gray box), \textbf{ReMD} recovers a compact, coherent vortex; image SR baselines smear or hallucinate textures, and ResShift-15 exhibits a softened core with ringing.
\item \textbf{Artifacts \& consistency.} Inside the dashed LR patch, \textbf{ReMD} refines details without checkerboard/ripples, while FNO/SwinIR show aliasing and EDSR banding—consistent with multigrid residual correction (Sec.~\ref{subsec:mgcorr}) and lightweight physics cues (Sec.~\ref{subsec:phys}).
\end{itemize}

\noindent\textit{Summary.} \textbf{ReMD-5} attains HR-like fronts and small eddies with fewer steps than ResShift-15, aligning with its spectral and RMSE gains.

\begin{table}[t]
\centering
\caption{\textbf{Ablation study (\textsc{Ocean}, $4\times$ SR).} Removing the MG corrector (“w/o residual correction”) or any physics residual (smoothing, anisotropic diffusion, spectrum) degrades all metrics; dropping the spectrum term yields the largest deterioration.}
\vspace{-8pt}
\label{table:ablation}
\begin{sc}
\resizebox{1.0\linewidth}{!}{
\begin{tabular}{l|ccc}
\toprule
Model Configuration & RMSE $\downarrow$ & PSNR $\uparrow$ & SSIM $\uparrow$ \\ 
\midrule
w/o Residual correction & 1.38E-02 & 47.37 & 0.981 \\
\midrule
w/o smoothing residual  & 1.40E-02 & 47.28 & 0.981 \\
w/o diffusion residual  & 1.40E-02 & 47.26 & 0.981 \\
w/o Spectrum residual   & 1.41E-02 & 47.20 & 0.981 \\
\midrule
ReMD (Baseline)         & \textbf{1.33E-02} & \textbf{47.71} & \textbf{0.983} \\ 
\bottomrule
\end{tabular}}
\end{sc}
\end{table}

\begin{table*}[t]
\centering
\caption{\textbf{Efficiency on the NS.} We report accuracy (RMSE$\downarrow$/PSNR$\uparrow$), parameter size, and wall-time (training time per epoch; inference time in seconds). Diffusion baselines (SR3, ResShift-15) incur high sampling cost; \textbf{ReMD-5} attains the best accuracy with fewer steps and lower inference time than ResShift, while \textbf{ReMD-2} is the fastest with competitive accuracy.}
\vspace{-8pt}
\label{table:efficiency}
\begin{sc}
\renewcommand{\multirowsetup}{\centering}
\resizebox{\linewidth}{!}{
\begin{tabular}{c|cccccc|cc|cc}
\toprule
\multirow{2}{*}{Metrics} & \multicolumn{10}{c}{Methods} \\ 
\cmidrule(lr){2-11} & EDSR & FNO & MWT & HiNOTE & Galerkin & SwinIR & SR3 & Resshift & \textbf{ReMD-2} & \textbf{ReMD-5} \\
\midrule
RMSE $\downarrow$ & 2.97E-02 & 4.42E-02 & 8.45E-02 & 8.03E-02 & 5.32E-02 & 3.63E-02 & 3.34E-01 & 2.21E-02 & 2.11E-02 & 2.09E-02 \\
PSNR $\uparrow$ & 46.89 & 43.44 & 37.81 & 38.25 & 41.83 & 45.15 & 25.87 & 49.47 & 49.84 & 49.94 \\
\midrule
Param Count & 1367553 & 684065 & 272605 & 420618 & 703777 & 3696633 & 93868033 & 112366821 & \multicolumn{2}{c}{118584941} \\
Param(MB) & 5.22 & 5.17 & 1.75 & 1.61 & 2.68 & 17.85 & 358.17 & 428.65 & \multicolumn{2}{c}{455.55} \\
\midrule
Training Time (s/epoch) & 8.87 & 4.78 & 1.72 & 57.91 & 146.25 & 86.26 & 53.48 & 64.1 & \multicolumn{2}{c}{100.33} \\
Inference Time (s/epoch) & 0.05 & 0.03 & 0.05 & 0.75 & 0.62 & 0.32 & 697.64 & 8.07 & 2.33 & 5.84 \\
\bottomrule
\end{tabular}}
\end{sc}
\end{table*}

\subsubsection{Ablation on multiscale correction and physics cues}
\label{subsec:ablation}
As summarized in Table~\ref{table:ablation}, removing any component degrades accuracy, with the multigrid corrector providing the main reduction of large-scale bias and the spectrum term contributing the largest share of high-frequency fidelity among the physics cues; smoothness and anisotropic diffusion act as stabilizers that suppress artifacts and preserve fronts. Overall, these parts are complementary—together they enable few-step sampling with the best performance.

\subsubsection{Efficiency}
\label{subsec:efficiency}
Compared with diffusion baselines (Table~\ref{table:efficiency}), \textbf{ReMD} is markedly more \emph{inference–efficient} at the same or better accuracy.
SR3 requires very long sampling (hundreds of steps), leading to orders–of–magnitude slower inference and much worse accuracy. ResShift shortens sampling to 15 steps and improves quality, but still trails ReMD both in error and speed. With only \emph{5} steps, \textbf{ReMD-5} attains the best RMSE/PSNR while running $\sim1.4\times$ faster than ResShift; with \emph{2} steps, \textbf{ReMD-2} remains more accurate than ResShift and is $\sim3.5\times$ faster.

The gains stem from treating each reverse step as a multigrid residual correction: fixed multiwavelet transfers and depthwise smoothers keep per–step cost $\mathcal{O}(HW)$ with a small constant, so wall–time scales primarily with the \emph{number of steps}. Although ReMD carries a slightly larger parameter footprint and higher per–epoch \emph{training} time (due to time–gated multiscale passes), inference is the dominant cost at deployment; on this Pareto front (error vs.\ time), ReMD strictly dominates both SR3 and ResShift.

\begin{table}[t]
  \centering
  \caption{\textbf{Physics-related metrics (lower is better $\downarrow$).}
  VE/EE are vorticity/enstrophy errors on NS2D ($\times4$). GED is the energy discrepancy on ERA5\_uo ($\times4$).
  The last column reports RMSE on ERA5\_uo $\times8$ super-resolution.}
  \label{tab:res}
  \vspace{-10pt}
  \footnotesize
  \setlength{\tabcolsep}{5.5pt}
  \renewcommand{\arraystretch}{1.05}
  \begin{tabular}{c|cc|c|c}
    \toprule
    Model & VE $\downarrow$ & EE $\downarrow$ & GED $\downarrow$ & RMSE ($\times8$) $\downarrow$ \\
    \midrule
    FNO         & 5.01E-03 & 2.57E-05 & 5.45E-02 & 3.76E-01 \\
    EDSR        & 6.43E-03 & 9.41E-05 & 4.38E-03 & 3.39E-01 \\
    SwinIR      & 6.16E-03 & 6.86E-05 & 5.10E-03 & 3.34E-01 \\
    LIIF        & 4.99E-03 & 2.08E-05 & 6.14E-03 & 3.32E-01 \\
    ResShift    & 3.26E-03 & 1.11E-05 & 4.51E-03 & 3.49E-01 \\
    \midrule
    \textbf{ReMD (ours)} & \textbf{2.34E-03} & \textbf{3.56E-06} & \textbf{4.24E-03} & \textbf{3.23E-01} \\
    \bottomrule
  \end{tabular}
\end{table}

\pgfplotsset{compat=1.18}

\begin{figure}[t]
  \centering
  \begin{tikzpicture}
    \begin{axis}[
      width=0.85\linewidth,
      height=0.4\linewidth,
      xlabel={Inference time (ms / sample)},
      ylabel={RMSE},
      ylabel style={yshift=-0.8em},
      grid=both,
      ymode=log,
      grid style={gray!20},
      legend cell align=left,
      legend style={font=\footnotesize, fill=white, draw=gray!30},
      legend pos=north east,
      xmin=0, xmax=3000,
      ymin=0.078, ymax=0.12,
    ]

      \addplot+[
        ultra thick,
        color=red,
        mark=o,
        mark size=1.8pt
      ] coordinates {
        (59.0245,  0.0804678)
        (130.3136, 0.0802478)
        (342.4443, 0.0802207)
        (685.1767, 0.0802242)
        (1021.2509,0.0802216)
        (1395.5013,0.0802243)
        (2815.9855,0.0802231)
      };
      \addlegendentry{ReMD (Full / Ours)}

      \addplot+[
        ultra thick,
        color=blue,
        mark=square,
        mark size=1.6pt
      ] coordinates {
        (44.9154,  0.1157081)
        (110.5240, 0.0896346)
        (220.8472, 0.0882943)
        (333.0799, 0.0879770)
        (442.1858, 0.0878295)
        (880.6320, 0.0876368)
        (1324.9001,0.0875584)
        (2406.5026,0.0874364)
      };
      \addlegendentry{Resshift (Baseline)}

      \addplot+[
        thick,
        dashed,
        mark=triangle*,
        mark size=1.6pt
      ] coordinates {
        (56.9887, 0.0906679)
        (130.4042, 0.0868701)
        (329.6344, 0.0827923)
        (681.1883, 0.0826609)
        (982.8243, 0.0826415)
        (1376.3123,0.0826353)
        (2713.0099,0.0826125)
      };
      \addlegendentry{w/o Residual Correction}

      \addplot+[
        thick,
        dashed,
        mark=diamond*,
        mark size=1.6pt
      ] coordinates {
        (56.9887, 0.0866679)
        (124.6959, 0.0824293)
        (338.5911, 0.0818422)
        (648.3304, 0.0819237)
        (1003.7103,0.0819362)
        (1249.8555,0.0819406)
        (2495.5826,0.0819652)
      };
      \addlegendentry{w/o Physics (All)}

      \addplot+[
        thick,
        dashed,
        mark=star,
        mark size=2.0pt
      ] coordinates {
        (46.8197, 0.1011227)
        (135.0117, 0.0939164)
        (328.1269, 0.0866911)
        (643.4340, 0.0862659)
        (1067.5766,0.0861716)
        (1392.8004,0.0861340)
        (2725.2233,0.0860363)
      };
      \addlegendentry{w/o Spectral}
    \end{axis}
  \end{tikzpicture}
  \vspace{-23pt}
  \caption{\textbf{Time--RMSE trade-off on ERA5\_uo ($\times4$ SR).} Each point varies the sampling steps; lower-left is better.}
  \label{fig:rmse_time}
\end{figure}
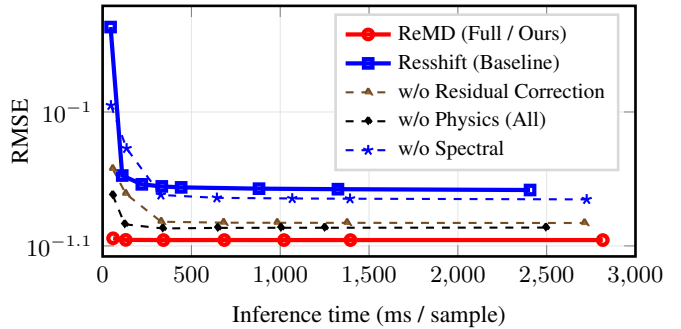

\begin{figure}[t]
  \centering
  \begin{tikzpicture}

    \begin{axis}[
      name=main,
      width=0.85\linewidth,
      height=0.38\linewidth,
      xmode=log,
      ymode=log,
      xlabel={Wavenumber $k$},
      ylabel={Energy spectrum $E(k)$},
      xmin=0.9, xmax=33,
      ymin=6e-7, ymax=0.5,
      grid=both,
      grid style={line width=0.1pt, draw=gray!25},
      major grid style={line width=0.2pt, draw=gray!35},
      legend style={
        at={(0.02,0.05)},
        anchor=south west,
        draw=none,
        fill=white,
        fill opacity=0.9,
        text opacity=1,
        font=\footnotesize,
        legend columns=2,
      },
      legend cell align=left,
      mark size=1.5pt,
      set layers,
    ]

      \addplot+[mark=none, very thick, color=blue] coordinates {
        (1,2.66939315e-01)
        (2,2.17743747e-02)
        (3,8.96609200e-03)
        (4,3.68482595e-03)
        (5,2.49021326e-03)
        (6,1.63633144e-03)
        (7,1.05446761e-03)
        (8,7.83106212e-04)
        (9,5.82713213e-04)
        (10,4.52179687e-04)
        (11,3.30261167e-04)
        (12,2.68242218e-04)
        (13,2.17961509e-04)
        (14,1.67453147e-04)
        (15,1.36019799e-04)
        (16,1.11968963e-04)
        (17,8.85129361e-05)
        (18,7.59297003e-05)
        (19,6.27465204e-05)
        (20,5.39556063e-05)
        (21,4.76912006e-05)
        (22,4.06343189e-05)
        (23,3.63998097e-05)
        (24,3.22297972e-05)
        (25,2.88406000e-05)
        (26,2.53283570e-05)
        (27,2.24756682e-05)
        (28,2.08098376e-05)
        (29,1.92161770e-05)
        (30,1.82063467e-05)
        (31,1.75223423e-05)
        (32,1.73518690e-05)
      };
      \addlegendentry{GT}

      \addplot+[mark=none, thick, color=green!70!black, dashdotted] coordinates {
        (1,2.66712438e-01)
        (2,2.16959384e-02)
        (3,8.89923372e-03)
        (4,3.62675779e-03)
        (5,2.42366984e-03)
        (6,1.57113462e-03)
        (7,9.91728004e-04)
        (8,7.01823378e-04)
        (9,5.02577593e-04)
        (10,3.77633368e-04)
        (11,2.64806969e-04)
        (12,2.02795930e-04)
        (13,1.55861666e-04)
        (14,1.12143043e-04)
        (15,8.50247629e-05)
        (16,6.58681732e-05)
        (17,4.82995691e-05)
        (18,3.75450200e-05)
        (19,2.88901042e-05)
        (20,2.09158231e-05)
        (21,1.53923765e-05)
        (22,1.04818927e-05)
        (23,6.68146625e-06)
        (24,4.34039274e-06)
        (25,2.64229272e-06)
        (26,1.85800860e-06)
        (27,1.35812656e-06)
        (28,1.12526770e-06)
        (29,1.12405701e-06)
        (30,1.31033912e-06)
        (31,2.46637987e-06)
        (32,2.07925886e-06)
      };
      \addlegendentry{SwinIR}

      \addplot+[mark=none, very thick, color=black, dashed] coordinates {
        (1,2.66532888e-01)
        (2,2.16038298e-02)
        (3,8.83915662e-03)
        (4,3.59622956e-03)
        (5,2.40063231e-03)
        (6,1.54683624e-03)
        (7,9.85556614e-04)
        (8,6.91805730e-04)
        (9,4.84342077e-04)
        (10,3.48028872e-04)
        (11,2.51334873e-04)
        (12,1.92974244e-04)
        (13,1.43324187e-04)
        (14,1.07601999e-04)
        (15,8.07956771e-05)
        (16,6.42403534e-05)
        (17,4.63177409e-05)
        (18,3.53740487e-05)
        (19,2.68119180e-05)
        (20,1.96811056e-05)
        (21,1.44503828e-05)
        (22,1.01346309e-05)
        (23,6.43263447e-06)
        (24,4.25544417e-06)
        (25,2.57682558e-06)
        (26,1.79605699e-06)
        (27,1.26349914e-06)
        (28,1.02487954e-06)
        (29,9.70842516e-07)
        (30,9.17202690e-07)
        (31,9.37594170e-07)
        (32,9.21190111e-07)
      };
      \addlegendentry{EDSR}

      \addplot+[mark=none, thick, color=yellow!70!black, dashed] coordinates {
        (1,2.66995530e-01)
        (2,2.18067631e-02)
        (3,8.93430591e-03)
        (4,3.65380320e-03)
        (5,2.43495402e-03)
        (6,1.61134061e-03)
        (7,1.02769463e-03)
        (8,7.47974144e-04)
        (9,5.20716106e-04)
        (10,3.88330442e-04)
        (11,2.86944338e-04)
        (12,2.30515562e-04)
        (13,1.82106682e-04)
        (14,1.39332179e-04)
        (15,1.07527701e-04)
        (16,8.92703530e-05)
        (17,6.90300117e-05)
        (18,5.76448923e-05)
        (19,4.81758930e-05)
        (20,3.91735681e-05)
        (21,3.41019258e-05)
        (22,2.79383597e-05)
        (23,2.37030943e-05)
        (24,2.10814529e-05)
        (25,1.80948747e-05)
        (26,1.59801834e-05)
        (27,1.41269095e-05)
        (28,1.25325776e-05)
        (29,1.14405061e-05)
        (30,1.06769083e-05)
        (31,1.01677802e-05)
        (32,9.96504151e-06)
      };
      \addlegendentry{Resshift}

      \addplot+[mark=none, very thick, color=red, dashdotted] coordinates {
        (1,2.67078893e-01)
        (2,2.17962766e-02)
        (3,8.97314563e-03)
        (4,3.68314787e-03)
        (5,2.48112935e-03)
        (6,1.63482000e-03)
        (7,1.05246061e-03)
        (8,7.70145395e-04)
        (9,5.70393742e-04)
        (10,4.33161023e-04)
        (11,3.11744526e-04)
        (12,2.50855735e-04)
        (13,2.03312826e-04)
        (14,1.58995767e-04)
        (15,1.24800722e-04)
        (16,1.02189139e-04)
        (17,7.97475530e-05)
        (18,6.69676727e-05)
        (19,5.65988421e-05)
        (20,4.78053035e-05)
        (21,4.21911004e-05)
        (22,3.60780494e-05)
        (23,3.18570943e-05)
        (24,2.92194832e-05)
        (25,2.53180307e-05)
        (26,2.27705354e-05)
        (27,2.02258482e-05)
        (28,1.85343691e-05)
        (29,1.77003659e-05)
        (30,1.66521946e-05)
        (31,1.61199295e-05)
        (32,1.62692552e-05)
      };
      \addlegendentry{ReMD}

      \draw[gray!60, very thin] (axis cs:20,8e-06) rectangle (axis cs:32,6e-05);

    \end{axis}

    \begin{axis}[
      name=inset,
      at={(main.north east)},
      anchor=north east,
      xshift=-2.5mm,
      yshift=-1.6mm,
      width=0.32\linewidth,
      height=0.10\linewidth,
      xmode=log,
      ymode=log,
      xmin=20, xmax=32.2,
      ymin=8e-06, ymax=6e-05,
      grid=both,
      grid style={line width=0.1pt, draw=gray!25},
      major grid style={line width=0.2pt, draw=gray!35},
      ticklabel style={font=\scriptsize},
      label style={font=\scriptsize},
      axis background/.style={fill=white},
      xlabel={},
      ylabel={},
      legend style={
        at={(0.02,0.98)},
        anchor=north west,
        draw=none,
        fill=white,
        fill opacity=0.9,
        text opacity=1,
        font=\scriptsize,
      },
      legend cell align=left,
    ]

      \addplot+[mark=none, very thick, color=blue] coordinates {
        (20,5.39556063e-05)
        (21,4.76912006e-05)
        (22,4.06343189e-05)
        (23,3.63998097e-05)
        (24,3.22297972e-05)
        (25,2.88406000e-05)
        (26,2.53283570e-05)
        (27,2.24756682e-05)
        (28,2.08098376e-05)
        (29,1.92161770e-05)
        (30,1.82063467e-05)
        (31,1.75223423e-05)
        (32,1.73518690e-05)
      };

      \addplot+[mark=none, thick, color=yellow!70!black, dashed] coordinates {
        (20,3.91735681e-05)
        (21,3.41019258e-05)
        (22,2.79383597e-05)
        (23,2.37030943e-05)
        (24,2.10814529e-05)
        (25,1.80948747e-05)
        (26,1.59801834e-05)
        (27,1.41269095e-05)
        (28,1.25325776e-05)
        (29,1.14405061e-05)
        (30,1.06769083e-05)
        (31,1.01677802e-05)
        (32,9.96504151e-06)
      };

      \addplot+[mark=none, very thick, color=red, dashdotted] coordinates {
        (20,4.78053035e-05)
        (21,4.21911004e-05)
        (22,3.60780494e-05)
        (23,3.18570943e-05)
        (24,2.92194832e-05)
        (25,2.53180307e-05)
        (26,2.27705354e-05)
        (27,2.02258482e-05)
        (28,1.85343691e-05)
        (29,1.77003659e-05)
        (30,1.66521946e-05)
        (31,1.61199295e-05)
        (32,1.62692552e-05)
      };

    \end{axis}

  \end{tikzpicture}
  \vspace{-23pt}
  \caption{NS2D energy spectrum (log--log).}
  \vspace{-15pt}
  \label{fig:spectrum}
\end{figure}
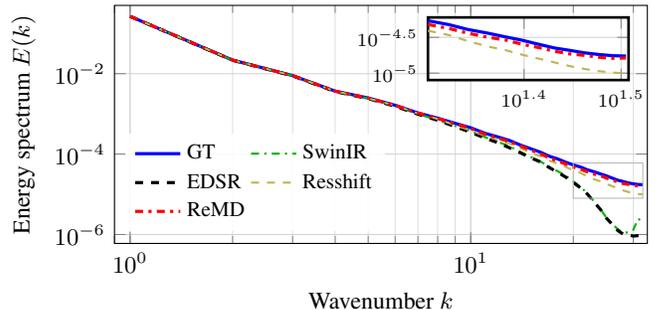

\section{Conclusion}
We presented \textbf{ReMD}, a physics-consistent diffusion framework that treats fluid SR as iterative residual correction. By coupling restriction consistency with lightweight physics cues and a time-gated multigrid corrector built on fixed multiwavelet transfers, ReMD attains state-of-the-art accuracy on NS, ERA5, and Ocean benchmarks with only 2–5 reverse steps, and produces spectrally faithful, low-divergence reconstructions.

\paragraph{Limitations \& future directions.}
(i) \emph{Pixel-space scaling.} Operating in pixel space constrains memory and compute at very high target resolutions; moving ReMD to latent domains can decouple cost from HR grid size and enable higher-resolution SR.
(ii) \emph{Mid-band accuracy.} Mid-frequency errors are reduced less effectively than low or high bands; introducing band-pass residuals, per-band (timestep-conditioned) gating, and refined spectrum-aware losses can better target the intermediate wavenumbers.
(iii) \emph{Temporal rollout.} Current evaluation is single-frame; integrating ReMD as a corrector in forecasting (predictor–corrector with neural operators), assessing long-horizon stability, and adding lightweight physics consistent projections during rollout address this gap.


{
    \small
    \bibliographystyle{ieeenat_fullname}
    \bibliography{main}
}


\end{document}